\documentclass[10pt,twocolumn,letterpaper]{article}

\usepackage[pagenumbers]{cvpr}

\usepackage[dvipsnames]{xcolor}

\usepackage{graphicx}
\usepackage{amsmath}
\usepackage{amssymb}
\usepackage{booktabs}
\usepackage{xcolor}
\usepackage{multirow} 
\usepackage{etoc}
\usepackage{minitoc}
\usepackage{bm}
\usepackage[accsupp]{axessibility}

\definecolor{cvprblue}{rgb}{0.21,0.49,0.74}
\usepackage[pagebackref,breaklinks,colorlinks,citecolor=cvprblue]{hyperref}

\def\themodel{Instruct 4D-to-4D\xspace} 

\makeatletter
\renewcommand\paragraph{\@startsection{paragraph}{4}{0mm}
                                   {0ex}
                                   {-1em}
                                   {\normalfont\normalsize\bfseries}}
\makeatother

\def\paperID{2769} 
\def\confName{CVPR}
\def\confYear{2024}

\title{\themodel: Editing 4D Scenes as Pseudo-3D Scenes Using 2D Diffusion }

\author{Linzhan Mou$^{1}$$^\dagger$ \qquad Jun-Kun Chen$^{2}$$^\dagger$ \qquad Yu-Xiong Wang$^{2}$ \vspace{0.1em} \\ 
    $^1$Zhejiang University \qquad $^2$University of Illinois Urbana-Champaign \qquad $^\dagger$Equal Contribution \vspace{0.1em}\\
    {\tt \hspace{0mm}moulz@zju.edu.cn \qquad{\tt \{junkun3,yxw\}@illinois.edu}}
}
\begin{document}

\twocolumn[{
\maketitle
\renewcommand\twocolumn[1][]{#1}
    \centering
    \vspace{-5.0mm}
    \includegraphics[width=1.0\linewidth]{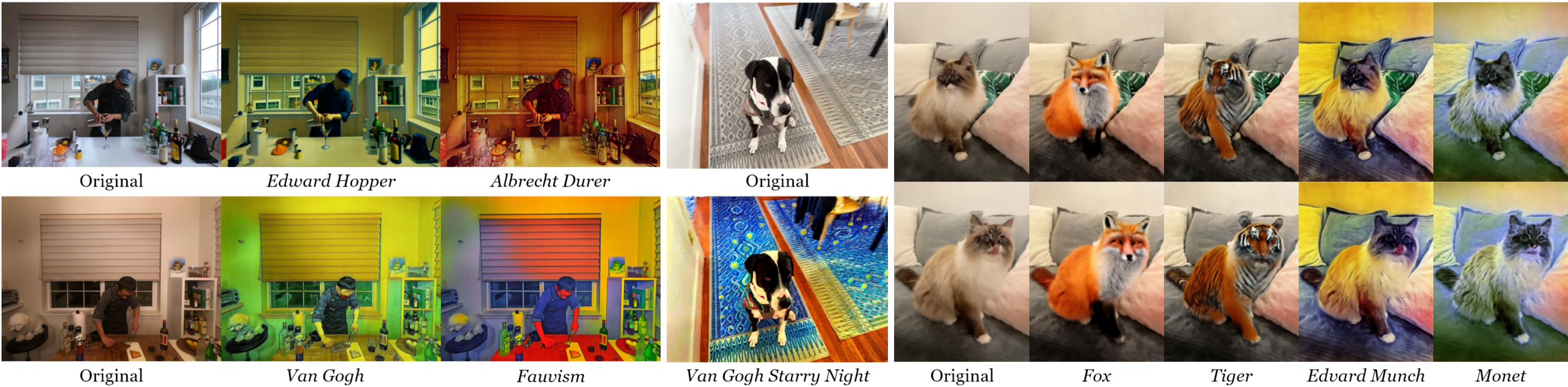}
    \captionof{figure}{\textbf{Our {\themodel} edits 4D scenes as pseudo-3D scenes with 2D diffusion}, achieving much sharper results with detailed textures across a variety of editing tasks and scenes. Notably, \themodel generates realistic and 4D consistent editing results in both monocular scenes and challenging multi-camera indoor scenes. \textbf{Please refer to the supplementary video for additional visualization}.} 
    \label{fig:teaser}
    \vspace{3mm}
}]

\begin{abstract}
\vspace{-4mm}
This paper proposes \themodel that achieves 4D awareness and spatial-temporal consistency for 2D diffusion models to generate high-quality instruction-guided dynamic scene editing results. Traditional applications of 2D diffusion models in dynamic scene editing often result in inconsistency, primarily due to their inherent frame-by-frame editing methodology. Addressing the complexities of extending instruction-guided editing to 4D, our key insight is to treat a 4D scene as a pseudo-3D scene, decoupled into two sub-problems: achieving temporal consistency in video editing and applying these edits to the pseudo-3D scene. Following this, we first enhance the Instruct-Pix2Pix (IP2P) model with an anchor-aware attention module for batch processing and consistent editing. Additionally, we integrate optical flow-guided appearance propagation in a sliding window fashion for more precise frame-to-frame editing and incorporate depth-based projection to manage the extensive data of pseudo-3D scenes, followed by iterative editing to achieve convergence. We extensively evaluate our approach in various scenes and editing instructions, and demonstrate that it achieves spatially and temporally consistent editing results, with significantly enhanced detail and sharpness over the prior art. Notably, \themodel is general and applicable to both monocular and challenging multi-camera scenes.
Code and more results are available at \href{https://immortalco.github.io/Instruct-4D-to-4D/}{\textit{immortalco.github.io/Instruct-4D-to-4D}}.

\end{abstract}
\vspace{-4mm}    
\section{Introduction}
\label{sec:intro}
Being able to synthesize photo-realistic novel-view images through rendering, neural radiance field (NeRF)~\cite{nerf} and its variants have become the leading neural representation for 3D and even 4D dynamic scenes. Moving beyond the mere representation of existing scenes, there is a growing interest in creating new, varied scenes sourced from an original scene via scene editing. The most convenient and straightforward way for users to communicate scene editing operations is through natural language -- a task known as instruction-guided editing. 

Success in this task for 2D images has been achieved by a 2D diffusion model, namely Instruct-Pix2Pix (IP2P)~\cite{instructpix2pix}. However, extending this capability to NeRF-represented 3D or 4D scenes poses a significant challenge. The inherent difficulty arises from the implicit nature of the NeRF representation, which lacks direct ways to modify the parameters in a targeted direction, along with the significantly increased complexity emerging in new dimensions. Recently, there has been noticeable progress in instruction-guided 3D scene editing, as exemplified by Instruct-NeRF2NeRF (IN2N) \cite{instruct-nerf2nerf}. IN2N achieves 3D editing through distillation from 2D diffusion models such as IP2P to edit NeRF, \ie, generating edited multi-view images from IP2P and fitting them on the NeRF-represented scenes. Due to the high diversity in generation results of diffusion models, IP2P may produce multi-view inconsistent images, with the same object having different appearances in different views. Therefore, IN2N consolidates the results by training on NeRF to make it converge to an ``average'' editing result, which is reasonable but often encounters challenges in practice.

Further extending the editing task from 3D to 4D, however, introduces fundamental difficulties. With the additional time dimension beyond 3D scenes, it requires not only 3D \emph{spatial consistency} for the 3D scene slice at each frame, but also the \emph{temporal consistency} between different frames. Notably, as recent 4D NeRFs~\cite{kplanes,nerfplayer} model the property of each absolute 3D location in the scene instead of the movement of individual object, the same object in different frames is not modeled by the same parameter. This deviation prevents NeRF from achieving spatial consistency by fitting inconsistent multi-view images, making the IN2N pipeline unable to effectively perform editing on 4D scenes.

This paper introduces \themodel, making the \emph{first} attempt in instruction-guided 4D scene editing that overcomes the aforementioned issues. \emph{Our key insight is to regard a 4D scene as a pseudo-3D scene}, where each pseudo-view is a video consisting of all frames from the same viewpoint. Subsequently, the task on the pseudo-3D scene can be tackled in a similar way as real 3D scenes, decoupled into two sub-problems: 1) achieve temporal-consistent editing for each pseudo-view, and 2) use the method from (1) to edit the pseudo-3D scene. Then, we can solve (1) with a video editing method, and leverage a distillation-guided 3D scene editing method to solve (2). 

Specifically, we utilize an \emph{anchor-aware attention} module to augment the IP2P model, inspired by~\cite{tune-a-video}. The ``anchor'' in our module is a pair of an image and its editing result as a reference for the IP2P generation. The augmented IP2P now supports batched input of multiple images, and the self-attention module in the IP2P pipeline is substituted with a cross-attention mechanism against the anchor image of this batch. Consequently, IP2P generates editing results based on the correlation between the current image and the anchor image, ensuring consistent editing within this batch. However, the attention module may not always correctly associate objects in different views, introducing potential inconsistency. 

To this end, we further propose an {\em optical flow-guided sliding window method} to facilitate video editing. Leveraging RAFT~\cite{raft}, we predict optical flow for each frame to establish pixel correspondence between two adjacent frames. This enables us to propagate editing results from one frame to the next, similar to a warping effect. With the augmented IP2P and optical flow, we can edit the video in temporal order, by segmenting frames and then applying editing to each segment while propagating the editing to the next segment. The process involves utilizing optical flow to initialize editing based on previous frames and subsequently applying the augmented IP2P with the last frame of the preceding segment serving as the anchor.

As a 4D scene contains a large number of frames at each view, it becomes time-consuming to compute all the views. To address this, we adopt a strategy inspired by ViCA-NeRF~\cite{vica} to edit pseudo-3D scenes based on key views. We first randomly select key pseudo-views and edit them using the aforementioned method. Then for each frame, we employ depth-based projection to warp the results from the key views to other views, and utilize weighted average to aggregate the appearance information, obtaining the editing results for all the frames. Given the complexity of 4D scenes, we apply the iterative editing procedure of IN2N to iteratively generate edited frames and fit the NeRF on the edited frames, until the scene converges. 

We conduct extensive experiments on both \emph{monocular} and \emph{multi-camera} dynamic scenes to validate the effectiveness of our approach. The evaluation shows the remarkable capabilities of our approach in both achieving sharper rendering results with significantly enhanced detail and ensuring spatial-temporal consistency in 4D editing (Fig.~\ref{fig:teaser}).

\textbf{Our contribution} is three-fold. (1) We introduce \themodel, a simple yet effective framework to perform instruction-guided 4D editing, by editing 4D scenes as pseudo-3D scenes via distillation from 2D diffusion models. (2) We propose the anchor-aware IP2P and the optical flow-guided sliding window method, enabling efficient and consistent editing of long videos or pseudo-views of any length. (3) With the proposed method, we develop a pipeline to iteratively generate fully and consistently edited datasets, achieving high-quality 4D scene editing in various tasks. Our work represents the first effort to investigate and address the general instruction-guided 4D scene editing, laying the foundation for this promising task.
\begin{figure*}[t!]
    \centering
    \vspace{-3mm}
    \centerline{\includegraphics[width=1\linewidth]{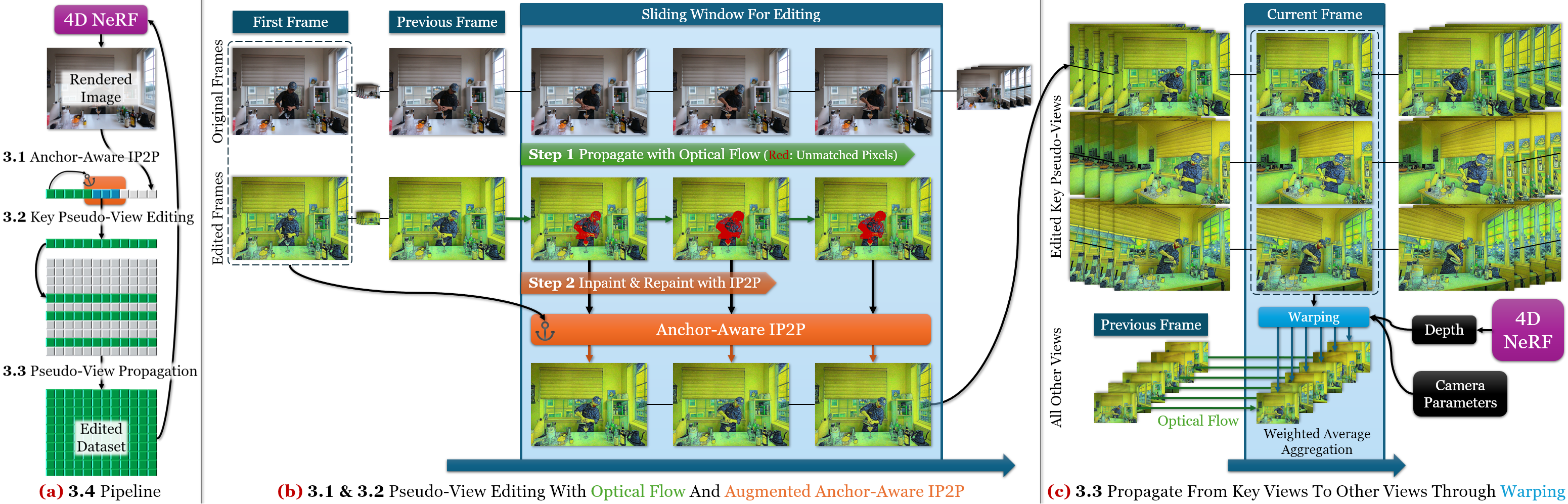}}
    \vspace{-3mm}
    \caption{Our \themodel edits a 4D scene by regarding it as a pseudo-3D scene with multiple pseudo-views, and then editing these pseudo-views in an iterative key frame-based pipeline. {\textcolor{red}{(a)}} Our pipeline edits the 4D scene by iteratively generating a fully edited dataset used to fit 4D NeRF. In each iteration, we first {\textcolor{red}{(b)}} edit each key pseudo-view through optical flow propagation and IP2P inpainting and repainting, and then {\textcolor{red}{(c)}} edit other pseudo-views by aggregating propagated results from both previous frames through optical flow, and the key pseudo-views at current frame through depth-based warping. 
    }
    \vspace{-6mm}
    \label{fig:pipeline}
\end{figure*}

\section{Related Work}

\paragraph{Diffusion-Based Video Editing.}

The diffusion-based generative models have achieved remarkable success in text-based image editing~\cite{dreambooth, diffedit, plug-and-play, p2p, sdedit, pix2pix-zero, instructpix2pix}. However, extending these models to video editing introduces greater complexity, necessitating the manipulation of visual attributes while maintaining temporal consistency. A prevalent approach in video editing using diffusion models is the transformation of Text-to-Image (T2I) models into Text-to-Video (T2V) models. Tune-A-Video~\cite{tune-a-video} incorporates temporal self-attention layers into UNet and performs the one-shot tuning. Make-A-Video~\cite{make-a-video} and MagicVideo~\cite{magicvideo} augment their networks by introducing the spatio-temporal attention (ST-Attn) mechanism, enabling the seamless transition of a pre-trained Text-to-Image model to the temporal dimension. Further, there is a growing focus on localized editing through the manipulation of attention maps inspired by Prompt-to-Prompt~\cite{p2p} and Plug-and-Play~\cite{plug-and-play}. Video-P2P~\cite{video-p2p} introduces decoupled-guidance attention control to preserve the semantic consistency. Pix2Video~\cite{pix2video} utilizes the self-attention feature injection to propagate modifications made in the anchor frame to other frames. Fatezero~\cite{fatezero} fuses self-attentions with a blending mask extracted from cross-attention features to achieve the zero-shot shape-aware editing.  

\paragraph{Diffusion-Based NeRF Editing.}
Diffusion-based NeRF editing has been gaining increasing attention in recent times. Some works leverage powerful SD as 2D prior to modifying the appearance of scenes, producing impressive results. Instruct 3D-to-3D~\cite{instruct-3dto3d} and Instruct-NeRF2NeRF (IN2N)~\cite{instruct-nerf2nerf} employ Instruct-Pix2Pix (IP2P)~\cite{instructpix2pix}, an image-conditioned diffusion model, to enable instruction-based 2D image editing. Specifically, Instruct 3D-to-3D uses score distillation sampling (SDS)~\cite{dreamfusion} loss to edit 3D NeRFs using the 2D diffusion-prior. Meanwhile, Instruct-NeRF2NeRF proposes Iterative Dataset Update (Iterative DU) to alternate between editing the images rendered from NeRF using the diffusion model and updating the NeRF representation with the supervision of edited images during the training process. ViCA-NeRF~\cite{vica} follows IN2N and utilizes the depth information derived from the NeRF to propagate the modification in key views to other views, achieving spatial consistency. DreamEditor~\cite{dreameditor} leverages DreamBooth~\cite{dreambooth} as 2D prior and utilizes the SDS loss to optimize the meshed-based neural field, performing faithful editing to the text. Control4D~\cite{control4d} proposes to build a more continuous 4D space by learning a 4D GAN~\cite{gan} from the ControlNet~\cite{controlnet} to avoid inconsistent supervision signals for 4D portrait editing.

\paragraph{NeRF-Based Dynamic Scene Representation}

The field of representing dynamic scenes using Neural Radiance Fields (NeRFs)~\cite{nerf} has seen significant advancements, which are essential for various real-world applications. Various methods have been developed to extend the capabilities of NeRFs in capturing and rendering dynamic scenes. DNeRF~\cite{dnerf} and Nerfies~\cite{nerfies} employ individual MLPs to represent a deformation ﬁeld and a canonical ﬁeld for capturing complex scene changes over time. DyNeRF~\cite{neural3d} integrates time-conditioning into NeRFs using a set of compact latent codes. TiNeuVox~\cite{TiNeuVox} employs an explicit voxel grid to model temporal information. Additionally, NeuralBody~\cite{neuralbody} and~\cite{structured, humannerf, relightable} focus on acquiring precise dynamic human body motion information, building upon the SMPL~\cite{smpl} model. HexPlane~\cite{hexplane} and K-Planes~\cite{kplanes} propose a planar factorization to decompose 4D spatiotemporal volumes into six feature planes. NeRFPlayer~\cite{nerfplayer} decomposes the 4D space into static, deforming, and new areas based on their temporal characteristics. Despite these advancements, a common limitation across these methods is the lack of user-friendly editing capabilities for dynamic scenes. Users are currently unable to freely edit or modify these scenes, particularly in terms of following specific instructions. This limitation highlights an area for potential future research and development, where user interactivity and editing capabilities could be integrated into the dynamic scene representation models. Addressing this challenge would significantly enhance the practicality and applicability of NeRF-based dynamic scene representations.
\section{Method}
\label{sec:method}

We propose \themodel, a novel pipeline that edits 4D scenes by distilling from Instruct-Pix2Pix (IP2P) \cite{instructpix2pix}, a powerful 2D diffusion model that supports instruction-guided image editing. The basic idea of our method roots in ViCA-NeRF~\cite{vica}, a key view-based editing. By regarding the 4D scene as a pseudo-3D scene where each pseudo-view is a video of multiple frames, we apply the key view-based editing, broken down into two steps: key pseudo-view editing, and propagation from key pseudo-views to other views, as shown in Fig. \ref{fig:pipeline}. We propose several key components to enforce and achieve spatial and temporal consistency during these steps, generating 4D consistent editing results. 

\subsection{Anchor-Aware IP2P for Consistent Batched Generation}
\label{sec:anchorip2p}

\paragraph{Batch Generation with Pseudo-3D Convs.} The editing process for a pseudo-view can be regarded as editing a video. Therefore, we need to enforce temporal consistency when editing each frame. Inspired by previous work on video editing \cite{tune-a-video,video-p2p}, we edit a batch of images together in IP2P, and augment the UNet in IP2P to make the generation in consideration of the whole batch. We upgrade its $3\times 3$ 2D convolutional layers to $1\times 3\times 3$ 3D convolutional layers by reusing the original parameters of kernels. 

\paragraph{Anchor-Aware Attention Module.} Limited by the GPU memory, we cannot edit all frames of a pseudo-view all together in one batch, and need to separate the generation into multiple batches. Therefore, it is crucial to keep the consistency between batches. Following the idea of Tune-a-Video \cite{tune-a-video}, instead of generating the edited result of the new batch from scratch, we allow the model to reference an anchor frame shared across all the generation batches, with its original and edited version, to ``propagate'' the editing style from it to the new edited batch. By substituting the self-attention module in the IP2P with the cross-attention model against the anchor frame, we would be able to connect the same objects between the current image and the anchor image, and generate the new edited images by mimicking the anchor's style, to perpetuate the consistent editing style from the anchor. Notably, our usage of the anchor-attention IP2P is different from Tune-a-Video, which queries cross-attention between the anchor frame and the previous frame instead of the current frame. Our design also further facilitates the inpainting procedure in Sec. \ref{sec:slidewindow}, which also requires a focus on the existing part of the current frame.

\begin{figure*}[t!]
    \centering
    \vspace{-3mm}
    \centerline{\includegraphics[width=0.8\linewidth]{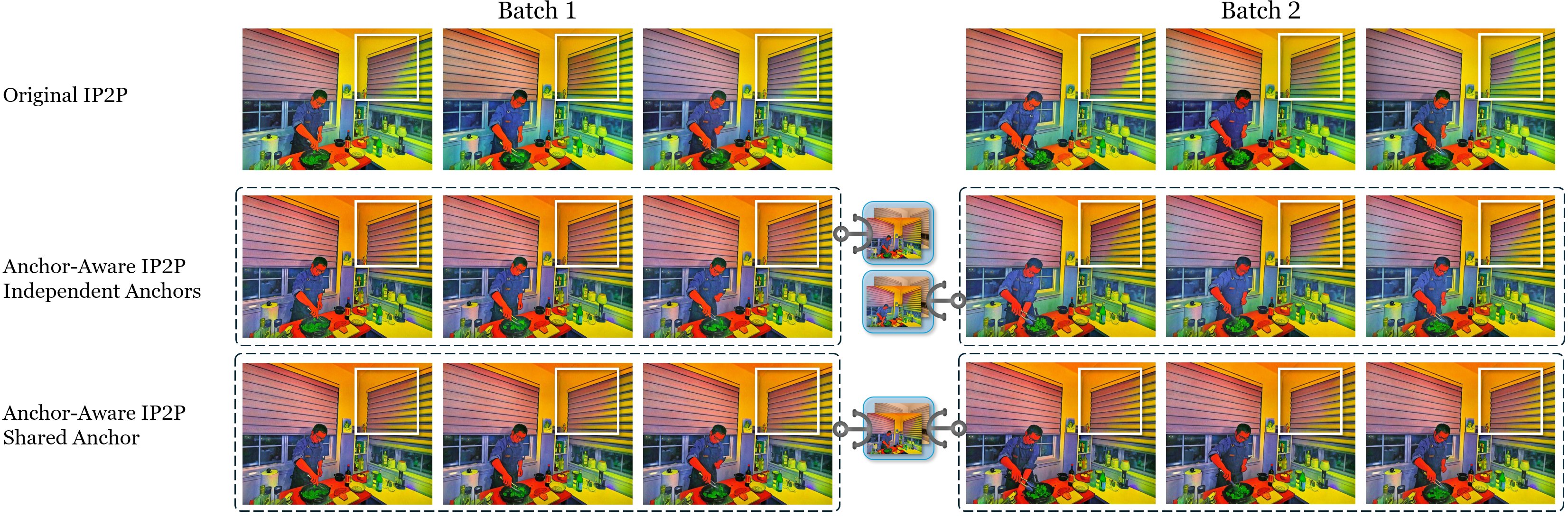}}
    \vspace{-3mm}
    \caption{Generation results show that our augmented IP2P achieves \emph{consistency within a batch} via our anchor-aware attention module, and achieves \emph{consistency between different batches} via the same anchor shared across batches. The white bounding box shows the most noticeable part of inconsistency.}
    \vspace{-6mm}
    \label{fig:exp-ip2p}
\end{figure*}

\paragraph{Effectiveness.}  Fig. \ref{fig:exp-ip2p} shows the generation results of different versions of IP2P. The original IP2P edits all images inconsistently, in different color distributions, even for images within one batch. With anchor-aware attention layers, IP2P is able to generate the batch as an entirety and, therefore, generates consistent editing results within one batch. However, it is still unable to generate consistent images within different batches. With reference to the same anchor image shared across batches, the full anchor-aware IP2P is able to generate consistent editing results for all 6 images across 2 batches, showing that even without additional training, the anchor-aware IP2P would be able to achieve consistent editing results.

\subsection{Optical Flow-Guided Sliding Window Method for Pseudo-View Editing}
\label{sec:slidewindow}
\paragraph{Optical Flow as 4D Warping. } To enforce the temporal consistency in a pseudo-view, we need the correspondence of the pixels between different frames. 3D scene editing methods \cite{csd,ivid,vica} exploits depth-based warping to find the correspondence of different views, using a deterministic method with NeRF-predicted depth and camera parameters. In 4D, however, there are no such explicit ways. Therefore, we use an optical flow estimation network RAFT \cite{raft} to predict the optical flow, in a format of 2D motion vector for each pixel, which can be derived into correspondence pixel in another frame. Using RAFT, we are able to warp between adjacent frames, like in 3D. As each pseudo-view is taken at a fixed camera location, optical flow performs well.
\paragraph{Sliding Window Method.} We follow the idea of video editing methods \cite{stylizing, interactive,tune-a-video,video-p2p} to edit a pseudo-view. However, those methods focus only on short videos and apply video editing by editing all the frames in a single batch, making them unable to deal with long videos. Therefore, we propose a novel sliding window (of width $B$ being the maximum allowed batch size) method to exploit the anchor-aware IP2P along with the optical flow. As shown in Fig. \ref{fig:pipeline} {\color{red} (a)}, for the current window containing $B$ images, say frames $t,t+1,\cdots,t+B-1$, we first propagate the editing results from frame $t-1$ to all these $B$ images one by one with 4D warping. For the unmatched pixels, which correspond to the occluded part in frame $t-1$, we set their value as their origin, obtaining a fused image of the original and warped editing images. 

Then, similar to the idea in ViCA-NeRF \cite{vica}, we use IP2P to inpaint and repaint the fused image at each view in the sliding window, by adding noise to the fused image and denoising using IP2P, so that the generated edited image will follow a similar pattern on the warped results, while repainting the whole image to make it natural and reasonable. To make the style all-consistent over the pseudo-view, we use the first frame as the anchor shared across all the windows, so that the model will generate images in a consistent style like the first view. As camera is fixed for one pseudo-view, there are many common objects between different frames, therefore such a method is very effective in producing consistent editing results for the frames in the window.

After completing the editing in the current window, we will advance the window by $B$ frames. Therefore, for a pseudo-view of $T$ frames in total, our \themodel only needs to call IP2P for $T/B$ times. By caching the optical flow prediction between adjacent frames in one view, we could achieve temporal-consistent pseudo-view editing very efficiently.

\subsection{Pseudo-View Propagation Based on Warping}
\label{sec:viewprop}

\paragraph{Generating First Frame.} As we need to propagate the edited pseudo-views to all other views while achieving spatial consistency, it is crucial to edit the first frames at all key pseudo-views in a spatially consistent way --  they are not only used to start the editing of the current pseudo-view, but also used as the anchor or the reference for all the proceeding generations. Therefore, we first edit one first frame in an arbitrary pseudo-key view, then use our anchor-aware IP2P with it as the anchor to generate other first frames. In this way, all the first frames are edited in a consistent style, being a good start to editing the key pseudo-views.

\paragraph{Propagate from Key Views to Other Views.}

After editing the key pseudo-views, aligned with ViCA-NeRF~\cite{vica}, we propagate their editing results to all other key views. ViCA-NeRF uses depth-based {\em spatial warping} to warp an image from another view at the same timestep, while we also propose optical-flow-based {\em temporal warping} to warp from the previous frame at the same view. With these two types of warping, we can warp the edited images from multiple sources.

We propagate for each timestep in the order of time. When we propagate at timestep $t$, for each frame $(v,t)$ at view $v$, we obtain its edited version as the weighted average of warped results from two sources: (1) the edited results of the previous frame at the same view, namely frame $(v,t-1)$, using temporal warping; and (2) the edited results of the current frame at one each of the key view, using spatial warping. By propagating the frames for all the timesteps, we obtain a consistent edited dataset containing all the editing frames. We use such a dataset to train NeRF towards the edited results.

With this propagation method, we would be able to efficiently generate a full dataset of consistently edited frames within $nT/B$ time-consuming IP2P generations, with $n$ key pseudo-views out of all $V$ pseudo-views, where in our experiments $n=5$ while $V$ can be more than $20$. Such high efficiency makes it possible to deploy an iterative pipeline to update the datasets.

\begin{figure*}[t!]
    \centering
    \vspace{-3mm}
    \centerline{\includegraphics[width=1.0\linewidth]{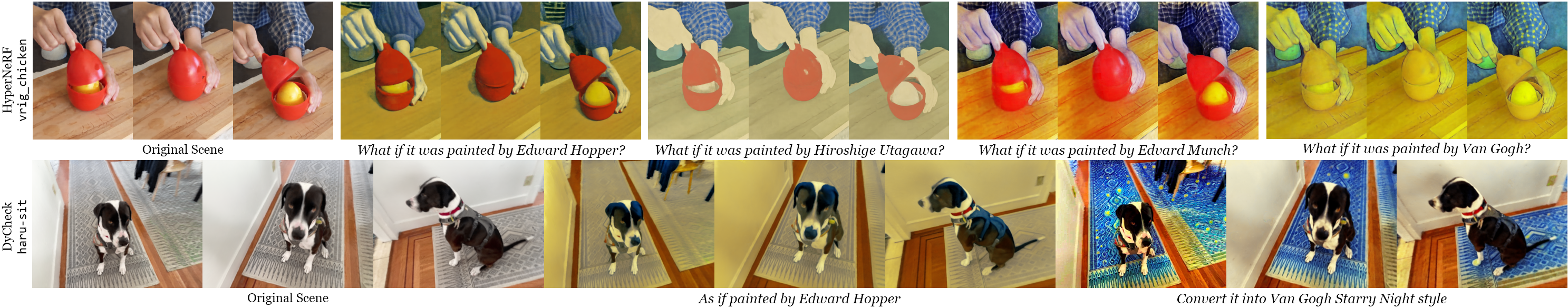}}   
    \centerline{\includegraphics[width=1.0\linewidth]{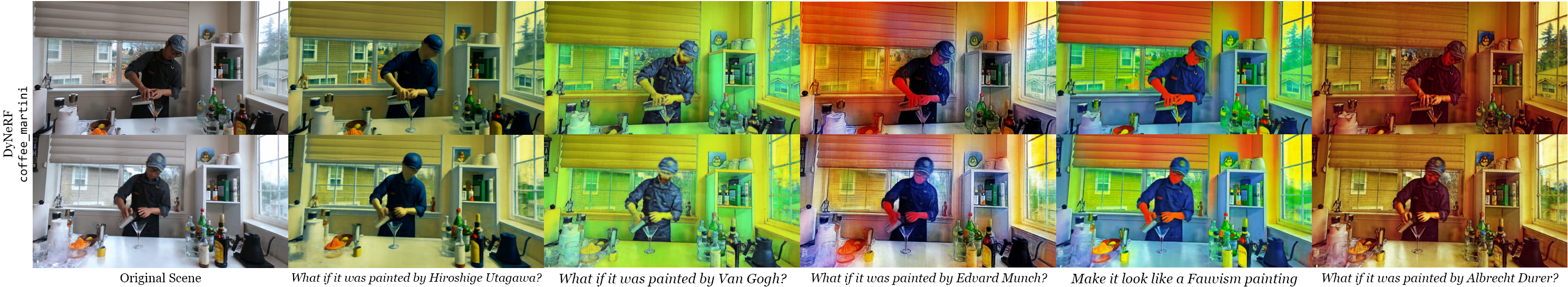}}   
    \centerline{\includegraphics[width=1.0\linewidth]{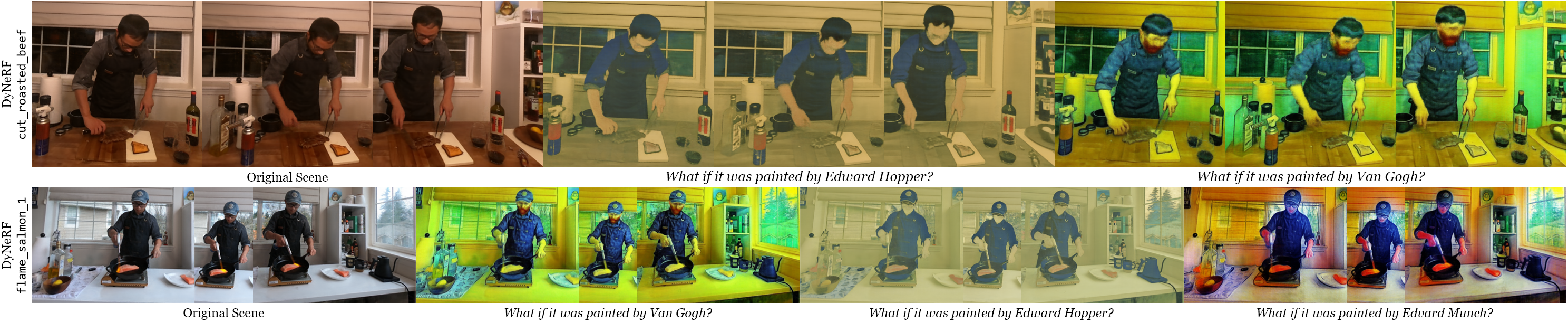}}   
    \vspace{-3mm}
    \caption{\textbf{Qualitative results on various scenes} demonstrate that our \themodel generates high-qualify editing results in style transfer tasks on various scenes. The edited scenes are well-consistent with the instructed style, showing bright colors and natural textures. }
    \vspace{-6mm}
    \label{fig:exp-gallery}
\end{figure*}

\subsection{Overall Editing Pipeline}

\paragraph{Iterative Dataset Update.}
Following the idea of IN2N~\cite{instruct-nerf2nerf}, we apply iterative dataset replacement on our baseline that iteratively re-generates the full dataset using the methods in Secs. \ref{sec:anchorip2p},\ref{sec:slidewindow},\ref{sec:viewprop}, and fits our NeRF on it. In each iteration, we first randomly select several pseudo-views as the key views in this generation. We use the method in Sec. \ref{sec:viewprop} to generate spatial-consistent editing results for the first frames of all these key pseudo-views, then propagate the editing for all pseudo-views using the sliding window method in Sec. \ref{sec:slidewindow}. After obtaining all the edited key pseudo-views, we use the method in Sec. \ref{sec:viewprop} to generate spatial and temporal consistent editing results for all other pseudo-views, ending up with a consistent edited dataset. We replace the 4D dataset with this edited dataset, and fit NeRF on it.

\paragraph{Improving Efficiency Through Parallelization and Annealing Strategies.}
In our pipeline, the NeRF only needs to be trained on the dataset and provide current rendering results, while IP2P only needs to generate results according to NeRF's rendering to form new datasets - there are few dependencies and interactions between IP2P and NeRF. Therefore, we parallelize our pipeline by running these two parts asynchronously on two GPUs. In the first GPU, we train NeRF continuously with the current dataset, while caching NeRF's rendering results in a rendering buffer; while in the second GPU, we apply our iterative dataset-generation pipeline to generate new datasets, using the images from the rendering buffer, and update the dataset used to train NeRF. In this case, we maximize the parallelization by minimizing the interactions, leading to a significant reduction in the training time.

On the other hand, to improve the generation results and convergence speed, we apply the annealing trick from HiFA~\cite{hifa} to achieve fine-grained editing on NeRF. The high-level idea is that we use the noise level to control the similarity of rendered results and IP2P's editing results. We generate the dataset at a high noise level to generate sufficiently edited results, and then gradually anneal the noise level to stick to the edited results that NeRF is converging to and refine such results. Instead of IN2N which always generates at a random noise level, our \themodel could converge to high-quality editing results at a fast speed.

With these two techniques, our \themodel is able to edit a large-scale 4D scene with 20 views and hundreds of frames in only hours.

\section{Experiments}
\label{sec:expr}

\begin{figure*}[t!]
    \centering
    \vspace{-3mm}
    \centerline{\includegraphics[width=0.9\linewidth]{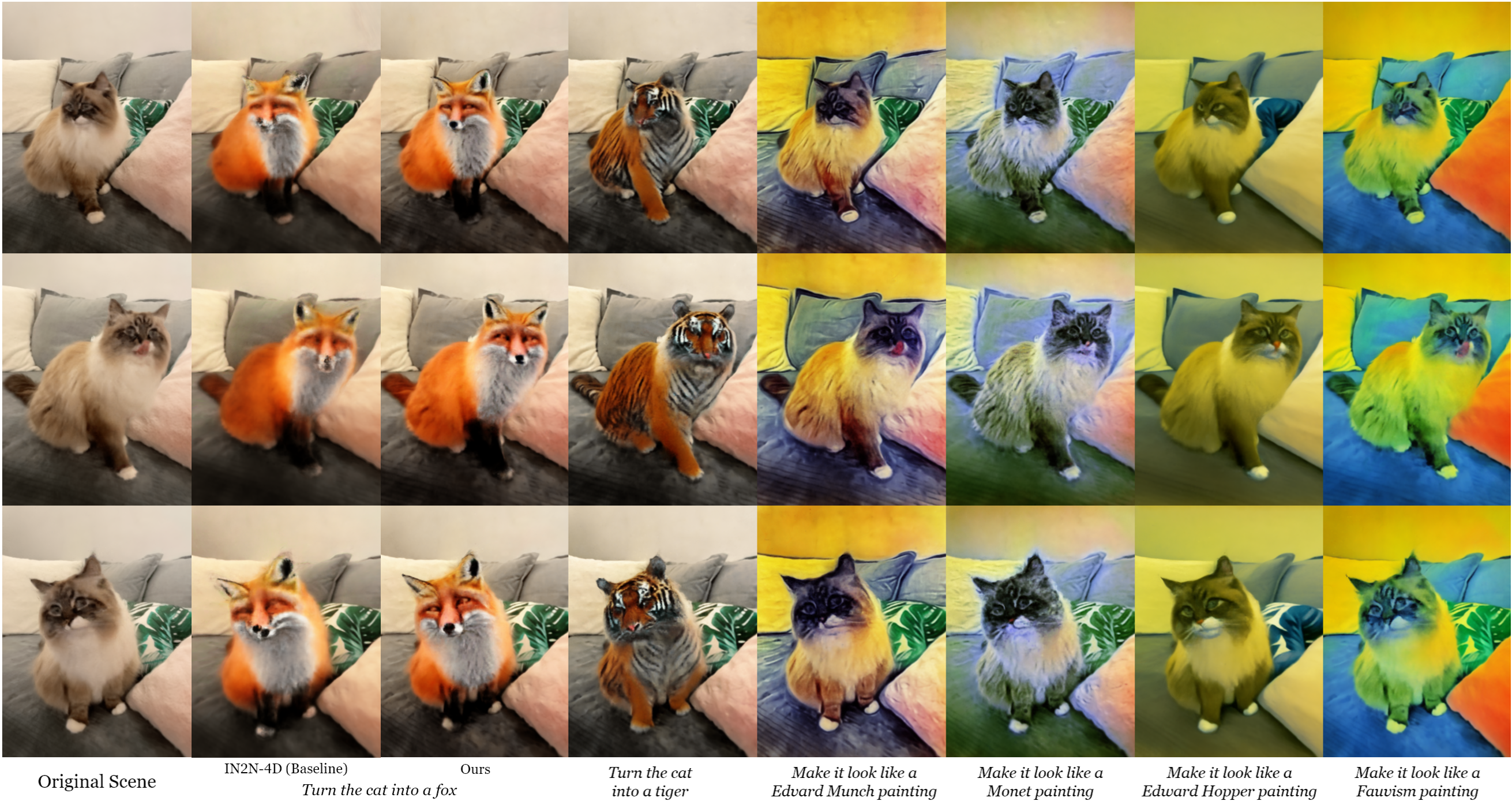}}
    \vspace{-3mm}
    \caption{\textbf{Qualitative results on \texttt{mochi-high-five} scene in DyCheck dataset} show that our \themodel achieves high-quality editing results over various editing instructions in the monocular scene. Our \themodel can even achieve consistent editing with complicated textures, \eg, in the Tiger editing, while baseline IN2N-4D generates blurred results with lots of artifacts. }
    \vspace{-3mm}
    \label{fig:exp-cat}
\end{figure*}

\begin{figure*}[t!]
    \centering

    \centerline{\includegraphics[width=0.9\linewidth]{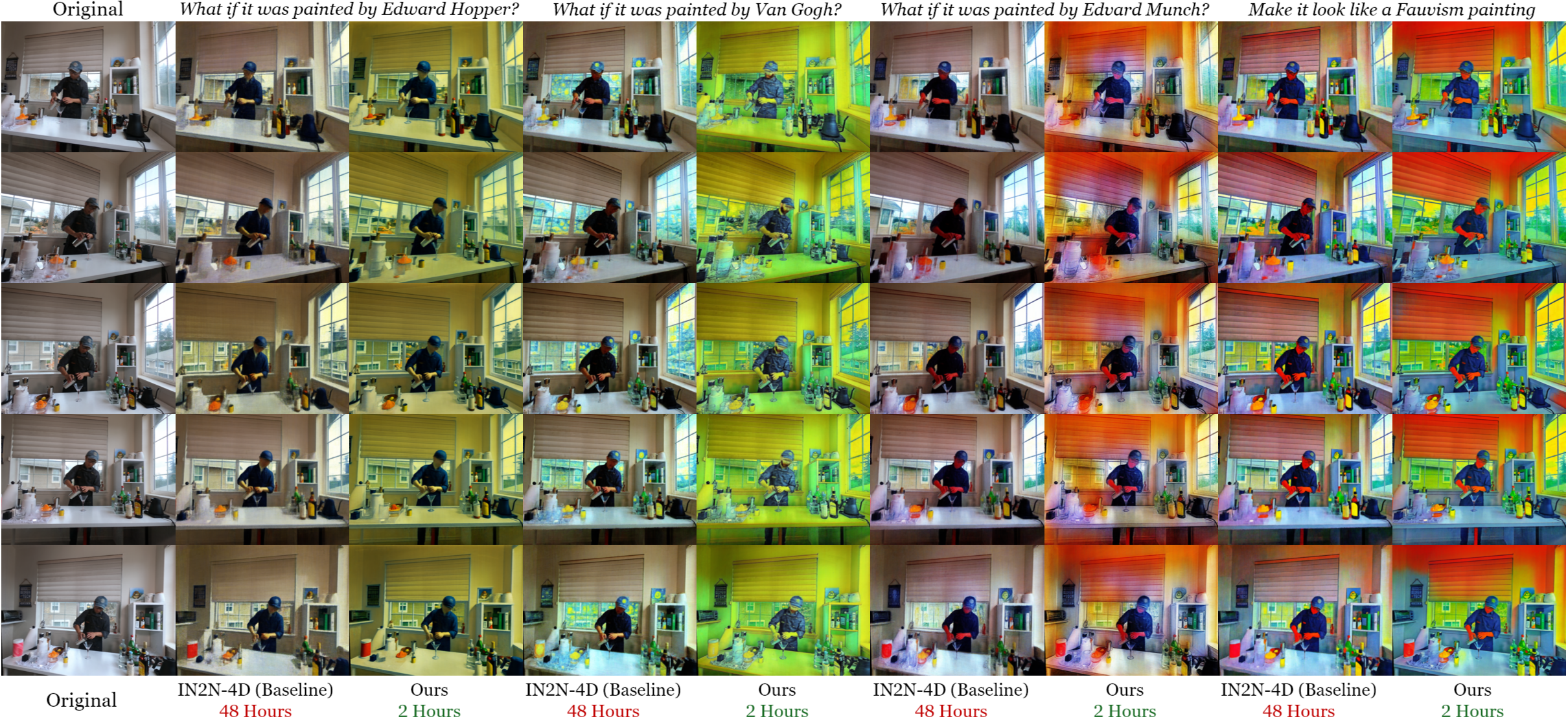}}
    \vspace{-3mm}
    
    \caption{\textbf{Qualitative comparison with baseline IN2N-4D} on multi-camera \texttt{coffee\_martini} shows that our \themodel generates high-quality, faithful style transfer editing results in a very short time.  As a comparison, IN2N-4D even fails to converge at any style with \textbf{24}$\times$ time consumption. }
    \vspace{-6mm}
    \label{fig:exp-compare-multi}
    
\end{figure*}

\paragraph{Editing Tasks and NeRF Backbone.}  The 4D scenes we use for evaluation are captured by single hand-held cameras and multi-camera arrays including: (I) {\em Monocular} Scenes in DyCheck \cite{dycheck} and HyperNeRF \cite{hypernerf}, which are simple, object-centric scenes with a single moving camera;  and (II) {\em Multi-camera} Scenes in DyNeRF/N3DV \cite{neural3d}, including indoor scenes with face-forward perspective and human motion structure. For monocular scenes, we edit all the frames as a single pseudo-view. We use the NeRFPlayer\cite{nerfplayer} as our NeRF backbone to produce high-quality rendering results of 4D scenes.

\paragraph{Baselines.} \themodel is the first work on instruction-guided 4D scene editing. No previous work focuses on the same task, while the only similar work Control4D~\cite{control4d} has not released their code. Therefore, we cannot conduct any baseline comparison with existing methods. To show the effectiveness of our \themodel, we construct a baseline IN2N-4D, by naively extending IN2N~\cite{instruct-nerf2nerf} to 4D, which iteratively generates one edited frame and add it to the dataset. We compare our \themodel with IN2N-4D both qualitatively and quantitatively. To quantify the results, as both our pipeline and the model are training NeRF with generated images, we use traditional NeRF~\cite{nerf} metrics to evaluate the results, namely PSNR, SSIM, and LPIPS, between the IP2P generated images (generating from pure noise so that it will not be conditioned on NeRF's rendering image) and the NeRF's rendering results. We conduct our ablation studies against \themodel variants in the supplementary.

\paragraph{Qualitative Results.} Our qualitative results are shown in Figs. \ref{fig:exp-compare-multi},~\ref{fig:exp-cat}, and~\ref{fig:exp-gallery}. The qualitative comparison with baseline IN2N-4D is in Figs. \ref{fig:exp-cat} and~\ref{fig:exp-compare-multi}. As shown in Fig. \ref{fig:exp-cat}, in the task of changing the cat into a fox in the monocular scene, IN2N-4D generates blur results with multiple artifacts: multiple ears, multiple noses and mouths, \etc, while our \themodel generates photo-realistic results where the shape of the fox is well aligned with the cat in the original scene, with clear textures on the fur and no artifacts. These results show that our anchor-aware IP2P, optical flow-based warping, and sliding window method for pseudo-view editing produces temporal-consistent editing results for a pseudo-view. Without such a module, the original IP2P in IN2N-4D produces inconsistent edited images for each frame, consolidating to a strange result on the 4D NeRF. Fig.~\ref{fig:exp-compare-multi} shows the style transfer results on multi-camera scenes. Our parallelized \themodel achieves consistent style transfer results that match the description in a very short time period of two hours, while IN2N-4D takes 24$\times$ longer than our \themodel but still fails to get the 4D NeRF converged to the indicated style. This shows that 4D scene editing is highly non-trivial, while our \themodel's strategy to iteratively generate a full edited dataset facilitates high-efficiency editing. All these results collectively show that all our design of \themodel is reasonable and effective, and \themodel can produce high-quality editing results in a very efficient way.

The experiments in Fig. \ref{fig:exp-cat} show the monocular scene \texttt{mochi-high-five} under different instructions, including local editing on the cat, or style transfer instructions for the whole scene. Our \themodel achieves photo-realistic local editing results in the Fox and Tiger instructions, with clear and consistent textures \eg, the stripes of the tiger. In the style transfer instructions, the edited scene faithfully reflects the style indicated in the prompts. These show \themodel's great ability in editing monocular scenes under various prompts.

The experiments in Fig. \ref{fig:exp-gallery} show other style transfer results, including monocular scenes in HyperNeRF and DyCheck and multi-camera scenes in DyNeRF. \themodel consistently produces high-fidelity style transfer results with bright colors and clear appearance in various styles.

\begin{table}[t]
\setlength{\tabcolsep}{2pt}
\centering
\scalebox{0.8}{\begin{tabular}{l|l|cccc}
 \hline\hline
 Instruction & Method & PSNR$\uparrow$ & SSIM$\uparrow$ & $\text{LPIPS}_{\text{Alex}}$$\downarrow$& $\text{LPIPS}_{\text{VGG}}$$\downarrow$ \\
 \hline 
\multirow{2}{*}{\em Van Gogh} & Ours & \textbf{23.62} & \textbf{0.820} & \textbf{0.220} & \textbf{0.349} \\
& IN2N-4D & 17.43 & 0.645 & 0.466 & 0.573 \\
\hline 
\multirow{2}{*}{\em Hopper} & Ours & \textbf{17.47} & \textbf{0.533} & \textbf{0.356} & \textbf{0.429} \\
& IN2N-4D & 11.96 & 0.299 & 0.655 & 0.645 \\
\hline 
\multirow{2}{*}{\em Munch} & Ours & \textbf{12.92} & \textbf{0.362} & \textbf{0.520} & \textbf{0.598} \\
& IN2N-4D & 11.96 & 0.299 & 0.655 & 0.645 \\
\hline 
\multirow{2}{*}{\em Fauvism} & Ours & \textbf{18.86} & \textbf{0.728} & \textbf{0.245} & \textbf{0.377} \\
& IN2N-4D & 14.15 & 0.520 & 0.408 & 0.532 \\
\hline 
\multirow{2}{*}{(Average)} & Ours & \textbf{19.67} & \textbf{0.635} & \textbf{0.323} & \textbf{0.405} \\
& IN2N-4D & 14.11 & 0.457 & 0.512 & 0.587 \\
 \hline\hline
\end{tabular}}
\vspace{-3mm}
\caption{In the quantitative evaluation on the multi-camera \texttt{coffee\_martini} scene, our \themodel significantly and consistently outperforms the baseline IN2N-4D in all metrics.  }
\vspace{-6.5mm}
\label{tab:exp-compare}
\end{table}

\paragraph{Quantitative Comparison.} The quantitative comparison between our \themodel and baseline IN2N-4D on the multi-camera \texttt{coffee\_martini} scene is in Tab. \ref{tab:exp-compare}. Consistent with the qualitative comparison, our \themodel significantly and consistently outperforms the baseline IN2N-4D. This shows that the NeRF trained by \themodel fits the IP2P's editing results much better than the baseline, further validating the effectiveness of our \themodel.

\paragraph{Ablation Study: Variants and Settings.}
We validate our design choices by comparing our approach to the following variants.
\begin{itemize}
    \item \textit{Video Editing.} This variant serves as the most basic implementation of our \themodel{} -- edit each pseudo-frame with any video editing methods, and propagate to other frames using 3D warping. We use a zero-shot text-driven video editing model, FateZero~\cite{fatezero}, via pretrained stable diffusion models. We follow the settings of the official Fatezero implementation in style editing and attribute editing. Since they can only process 8 video frames in a batch, we use a batch-by-batch editing strategy for pseudo-view editing. 
    \item \textit{Anchor-Aware IP2P w/o Optical-Flow.} In this variant, we perform video editing without optical flow guidance, in which anchor-aware IP2P directly edits all training images using the same diffusion model setting.
    \item \textit{One-time Pseudo-View Propagation.} In this variant, we perform
    only one-time pseudo-view propagation, in which all rest-pseudo-views are warped from 4 randomly selected key-pseudo-views, and the NeRF is trained until convergence on those edited images. 
\end{itemize}

The task for ablation study is ``What if it was painted by Van Gogh'' on the \texttt{coffee\_martini} in DyNeRF dataset. As the ``Video Editing'' variant does not use the same diffusion model, IP2P \cite{instructpix2pix}, to edit the video, we cannot use the metrics in the main paper. Therefore, consistent with IN2N \cite{instruct-nerf2nerf}, we use CLIP \cite{CLIP} similarity to evaluate how successful an editing operation is applied.
    
\paragraph{Ablation Study: Results.}

\begin{figure}[t]
    \centering
    
    \centerline{\includegraphics[width=1\linewidth]{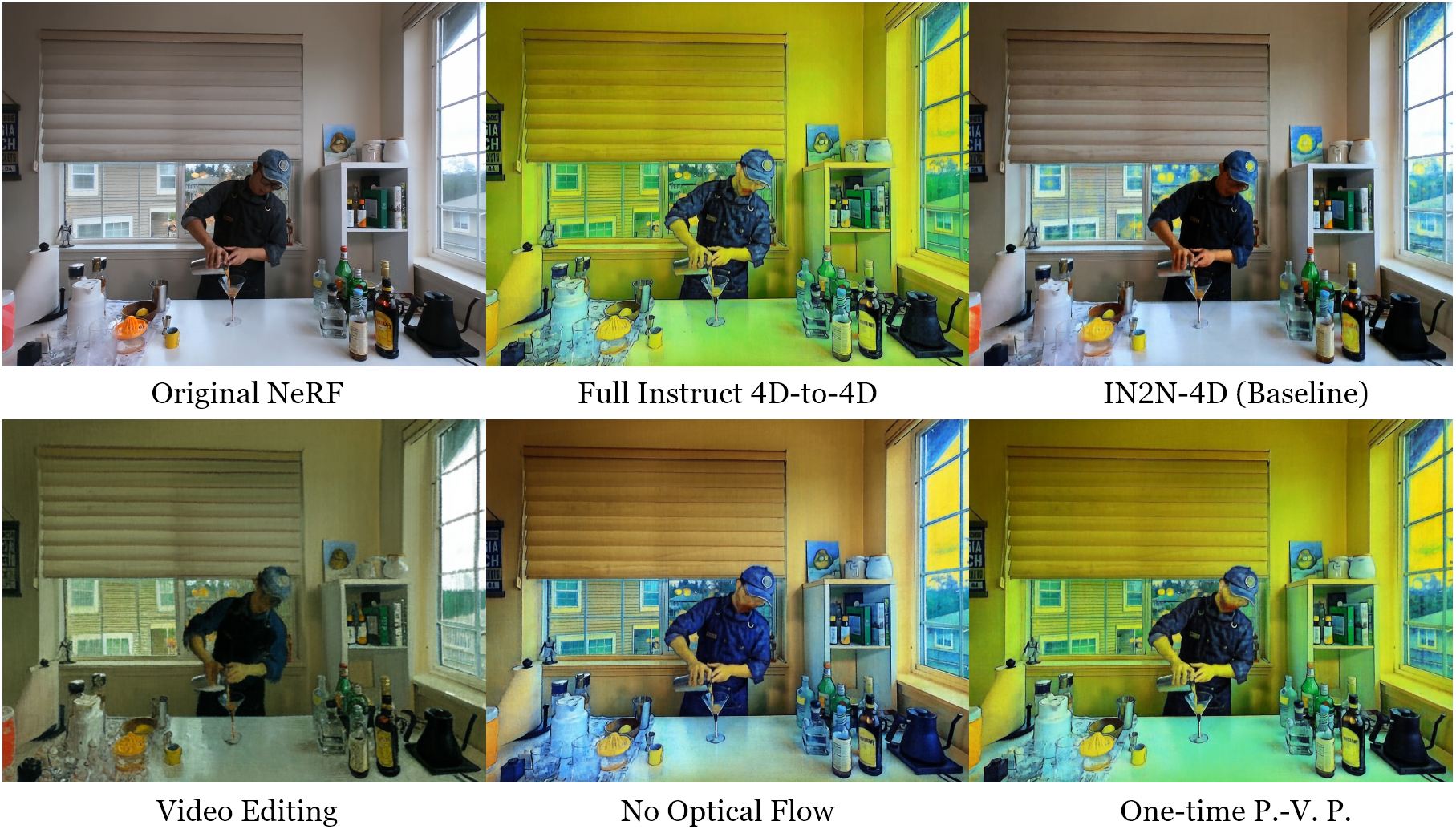}}
    \vspace{-3mm}
    \caption{Most of the variants are unable to edit the scene sufficiently, showing that the design choices of our \themodel are reasonable and effective. These results are also shown in the demo video. }
    \vspace{-3mm}
    \label{fig:exp-suppl-abla}
\end{figure}

\begin{table}[t]
\setlength{\tabcolsep}{2pt}
\centering
\begin{tabular}{l|l|c}
        \hline Type & Name & 
             CLIP Similarity$\uparrow$ \\
        \hline 
        Baselines & IN2N-4D  & 0.2790  \\
        \hline 
       \multirow{3}{*}{Our Variants} &  Video Editing  & 0.2631 \\
         & Ours w/o Optical Flow  & 0.2792 \\
         & One-time P.-V. P. & 0.2876 \\
         \hline
          & Full (Ours) & \textbf{0.3085} \\
        \hline
    \end{tabular}
\vspace{-3mm}
\caption{Ablation study showing the effectiveness of our design choices. The optical flow and iterative editing pipeline help achieve successful editing. }
\vspace{-6.5mm}
\label{tab:exp-ablation}
\end{table}

The qualitative results are in Fig. \ref{fig:exp-suppl-abla} and the demo video. Most of the variants do not apply sufficient editing to the scene, with a gloomy appearance and no Van Gogh's representative color. This shows that our \themodel's design choices are effective and crucial to achieve high-quality editing. 

The quantitative comparisons are in Tab. \ref{tab:exp-ablation}. Our full \themodel achieves significantly higher CLIP similarity in the ablation task, showing that our design is effective. Also, we observe that the Video Editing strategy cannot even achieve a better metric than IN2N-4D, which shows that it is non-trivial to edit the 4D scene even after converting it to a pseudo-3D scene.

\section{Conclusion}

This paper proposes \themodel, the first instruction-guided 4D scene editing framework that edits 4D scenes by regarding them as pseudo-3D scenes and applies an iterative strategy to edit pseudo-3D scenes using a 2D diffusion model. Qualitative experimental results show that \themodel achieves high-quality editing results in various tasks, including monocular and multi-camera scenes. \themodel also significantly outperforms the baseline, a naive extension of the state-of-the-art 3D editing method to 4D, showing the difficulty and non-trivialness of the task and the success of our method. We hope that our work could inspire more future work on 4D scene editing.

\noindent{\footnotesize\textbf{Acknowledgement.} This work was supported in part by NSF Grant 2106825, NIFA Award 2020-67021-32799, the Jump ARCHES endowment, and the IBM-Illinois Discovery Accelerator Institute. This work used NVIDIA GPUs at NCSA Delta through allocations CIS220014 and CIS230012 from the ACCESS program.}

{
    \small
    \bibliographystyle{ieeenat_fullname}
    \bibliography{main}
}

\twocolumn[{
\maketitle
\renewcommand\twocolumn[1][]{#1}
    \centering
    \Large
    \vspace{-1em}\textbf{Supplementary Material} \\
    \vspace{1.0em}
}]

\appendix

Our supplementary material includes implementation details and some additional experiments.
\renewcommand{\thetable}{\Alph{section}.\arabic{subsection}}
\renewcommand{\thefigure}{\Alph{section}.\arabic{subsection}}
\setcounter{tocdepth}{1}
\etocsetlevel{section}{1}
\def\authcount{}
\etocsettocstyle{\color{white}}{}
\renewcommand{\contentsname}{}
\localtableofcontents

\section{Demo Video \& Code}
We provide a demo video on our project page \href{https://immortalco.github.io/Instruct-4D-to-4D/}{\textit{immortalco.github.io/Instruct-4D-to-4D}}, visualizing our editing results as videos.

We also provide some source code in the \texttt{code} folder, which contains the code of our main pipeline.

\section{Implementation Details}

\subsection{4D Representation} 
The primary input of our method is a 4D NeRF representation, acquired through the NeRFPlayer~\cite{nerfplayer}. Our framework is general, and therefore, any 4D scene representation adopting RGB observations as supervision can be used. In the implementation of the multi-camera scenes, we use TensoRF~\cite{tensorf}-based NeRFPlayer as NeRF backbone and follow the same setting as in their experiments on the multi-camera DyNeRF~\cite{neural3d} scenes. For our experiments, we extract 50 frame segments, from the full-length videos and downsample images to $1352 \times 1014$ for 4D representation and editing. Furthermore, to show the capabilities of our method in long-term videos(pseudo scenes), we also use full-length 300-frame videos with $676\times 507$ resolution for additional evaluation, both trained for 100,000 iterations per scene. In the implementation of the monocular scenes, we use InstantNGP~\cite{instantngp}-based NeRFPlayer for HyperNeRF~\cite{hypernerf} dataset, trained for 60, 000 iterations per scene, and Nerfacto-based NeRFPlayer from NeRFStudio~\cite{nerfstudio} for DyCheck~\cite{dycheck} dataset, trained for 30, 000 iterations per scene.

\subsection{Anchor-Aware Attention}
We introduce the anchor-aware attention module to augment IP2P, enhancing the appearance control by the reference anchor frame. Specifically, each latent feature calculates the corresponding key and value features based on the concatenation of the anchor frame $z_{v_a}$ and the current frame $z_{v_i}$.  The definition of anchor-aware attention is as follows:
$$
    Q=W^Q z_{v_i}, K=W^K\left[z_{v_a} ; z_{v_i}\right], V=W^V\left[z_{v_a} ; z_{v_i}\right] 
$$
where $W$ are projection layers in attention shared across space and time, and $[\cdot]$ denotes concatenation operation. Besides, we employ the original spatial self-attention weights as initialization. In each pseudo-view editing process, we use the first frame which is edited as the anchor frame to provide appearance reference.

Fig. \ref{fig:suppl-impl}-(a) shows that anchor-aware attention clearly improves the editing consistency across views and batches.

\subsection{Sliding Window-Based Pseudo-View Editing Method}

\begin{figure}[t!]
    \centering
    
    \centerline{\includegraphics[width=1\linewidth]{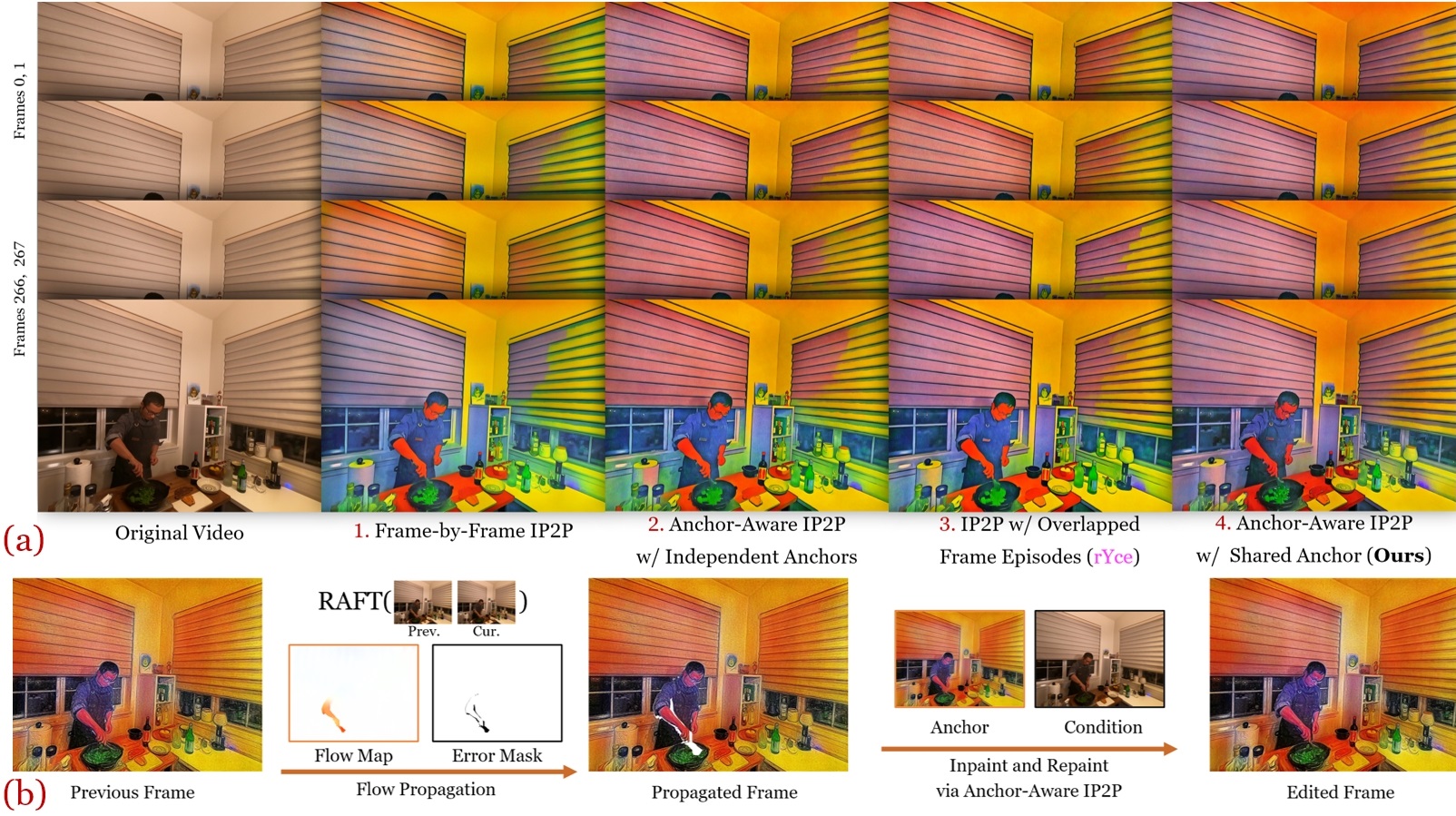}}
    \caption{(a) The usage of anchor-aware attention clearly improves the editing consistency across views and batches. (b) A more detailed visualization of our sliding window pseudo-view editing method through an optical flow-based propagation and anchor-aware inpainting and repainting. }
    \label{fig:suppl-impl}
\end{figure}

As visualized in Fig. \ref{fig:suppl-impl}-(b),  we filter inferior flow predictions by leveraging the forward-backward consistency constraint. Also, the anchor-aware IP2P is well designed both to inpaint occluded areas and to repaint the whole part based on the appearance of the anchor frame. This leads to reasonable editing results even when the RAFT prediction is not accurate.

\subsection{4D Editing Procedure}
We re-initialize the optimizers in the trained NeRFPlayer model and utilize Anchor-Aware Instruct-Pix2Pix~\cite{instructpix2pix} as our 2D editing model. For the diffusion model, varying hyperparameters are applied at distinct phases. During the anchor frames editing stage, the input diffusion timestep $t$ decays from 0.98 to 0.7 in a cosine annealing manner. We employ 20 diffusion steps for this phase. During the inpainting stage after optical flow warping, we set timestep $t$ to 0.6, and only used 3 diffusion steps. As optical flow warping extensively propagates across most areas, the diminished timestep and fewer diffusion steps contribute to efficient inpainting while preserving the original data distribution. To control the extent of alterations for specific edits, we calibrate the classifier-free guidance weights for each scene, defined as $S_T$ and $S_I$ for the text instruction and original image, respectively. For object-focused editing tasks, We set $S_I$= 1.5 and $S_T$ = 7.5, whereas, for style transfer tasks, the settings are $S_I$= 1.5 and $S_T$ = 9.5. The number of iterations varies in different scenes shown in the paper. Due to the parallelization strategy, we don't need to trade off between NeRF training and image editing, thus we use 15,000 iterations for monocular scenes and 25,000 iterations for multi-camera scenes (with over 1000 images each). While each experiment is conducted on 2 NVIDIA A40 GPUs, most scenes converge to a fine-grained edited scene within 1.5 hours.

\section{Discussion}
\subsection{Limitations} 
The major limitation of our \themodel is rooted in the limitation of IP2P~\cite{instructpix2pix} -- given that \themodel edits scenes by distilling from IP2P, its editing capability is capped by IP2P. We will fail in the failure cases of IP2P, and perform poorly if IP2P does so. In addition, as we are using the original IP2P without fine-tuning, we lose the ability to leverage the per-scene information to facilitate editing. On the other hand, we benefit from the high efficiency of such a training-free pipeline.

Moreover, without input of 3D geometry information or 4D movement information, IP2P is unaware of any 3D/4D information, including position, geometry, and timestep. It can only infer the correlation between frames using cross-attention modules based on the RGB images, which might be inaccurate and lead to inconsistent editing results. Note that the source of consistency in \themodel is primarily the cross-attention module, which is a soft mechanism without supervision or enforcement. While Fig. \ref{fig:exp-ip2p} shows that our IP2P can generate consistent editing results under certain situations, this is not always guaranteed.

Some instructions may indicate shape editing. \themodel could only perform simple shape editing where the modification is near the surface, \eg, `change the cat to a fox' which slightly changes the head shape. \themodel does not support aggressive shape editing, \eg, `remove the cat,' like most of the instruction-guided 3D scene editing methods, or editing the movement of an object.

\subsection{Future Directions}
One possible future direction is to support per-scene training, \eg, fine-tuning RAFT for more accurate optical flow prediction, augmenting IP2P to support 3D and 4D information, \etc This could lead to a more powerful IP2P towards more consistent 4D editing.

\end{document}